\documentclass[10pt,twocolumn,letterpaper]{article}
\pdfoutput=1
\usepackage{wacv}
\usepackage{times}
\usepackage{epsfig}
\usepackage{graphicx}
\usepackage{amsmath}
\usepackage{amssymb}
\usepackage{algorithm}
\usepackage{algorithmic}
\usepackage{enumerate}
\usepackage{dsfont}
\usepackage{comment}

\usepackage[pagebackref=true,breaklinks=true,letterpaper=true,colorlinks,bookmarks=false]{hyperref}

\wacvfinalcopy 

\ifwacvfinal\fi
\begin{document}

\title{Implicit Discriminator in Variational Autoencoder}

\author{Prateek Munjal\thanks{Authors with equal contribution}\\
Indian Institute of Technology\\
Ropar\\
{\tt\small 2017csm1009@iitrpr.ac.in}
\and
Akanksha Paul\footnotemark[1]\\
Indian Institute of Technology\\
Ropar\\
{\tt\small akanksha.paul@iitrpr.ac.in}
\and 
Naraynan C Krishnan\\
Indian Institute of Technology\\
Ropar\\
{\tt\small ckn@iitrpr.ac.in}
}

\maketitle
\ifwacvfinal\thispagestyle{empty}\fi


\begin{abstract}
Recently generative models have focused on combining the advantages of variational autoencoders (VAE) and generative adversarial networks (GAN) for good reconstruction and generative abilities. In this work we introduce a novel hybrid architecture, Implicit Discriminator in Variational Autoencoder (IDVAE), that combines a VAE and a GAN, which does not need an explicit discriminator network. The fundamental premise of the IDVAE architecture is that the encoder of a VAE and the discriminator of a GAN utilize common features and therefore can be trained as a shared network, while the decoder of the VAE and the generator of the GAN can be combined to learn a single network. This results in a simple two-tier architecture that has the properties of both a VAE and a GAN. The qualitative and quantitative experiments on real-world benchmark datasets demonstrates that IDVAE perform better than the state of the art hybrid approaches. We experimentally validate that IDVAE can be easily extended to work in a conditional setting and demonstrate its performance on complex datasets.
\end{abstract}

\section{Introduction}

Deep Variational Autoencoders(VAE\cite{VAE_kingma}) and Generative Adversarial Networks(GAN\cite{Goodfellow:2014:GAN:2969033.2969125}) are two recently used approaches in the generative modeling world. VAE is more stable in training but generates blurry samples. While GAN has the appealing property of generating realistic images; training a GAN is well known to be challenging leading to problems such as mode collapse.

Several recent approaches have proposed hybrid models of autoencoder and adversarial networks with a joint objective of achieving stable training like VAE and inferencing ability like GAN. In order to introduce the adversarial loss component in the objective functions most of the recent hybrid approaches include an adversary network that results in a three-tier architecture \ie an encoder, a decoder, and an adversary network. We hypothesize that the encoder and discriminator networks can share common layers encoder itself can be reused as a discriminator, thereby assuming an overlap in the knowledge learned by the encoder and the discriminator network. 

\begin{figure}[t]
\begin{center}
   \includegraphics[width=\linewidth]{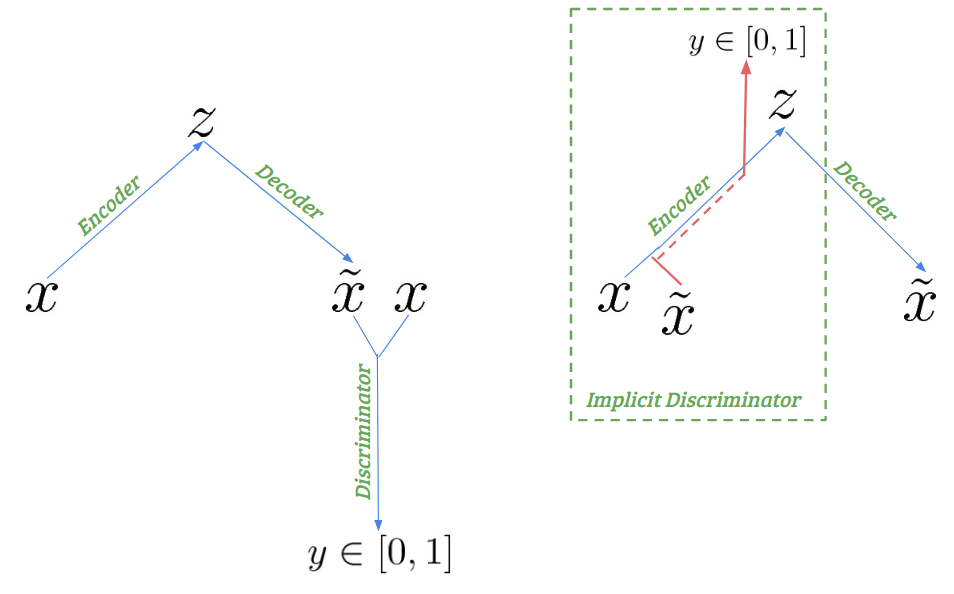}
\end{center}
\caption{Flow diagram of traditional hybrid approaches(left) and our proposed approach(right). We introduce the adversarial loss by collapsing the encoder into the discriminator, which we term as Implicit Discriminator. The output of discriminator is denoted by $y \in [0,1]$ where $0$ and ${1}$ represent fake and real respectively.}
\label{fig:approach_caption}
\label{fig:approach_onecol}
\end{figure}

The key idea behind our approach is that we would like the discriminator to provide the useful gradients to the generator if it misses any mode in the true data distribution. The traditional GAN learning does not explicitly encourage such a property in the discriminator and therefore, we suspect that it is 
vulnerable to the issue of mode collapse. Further, we note that the traditional L2 loss used for learning the encoder can be seen as minimizing the forward-KL divergence. The forward-KL divergence comes with the mode inclusive property (\ie it never misses a mode in the true data distribution). Therefore, to make the discriminator aware of all the modes in the true data we propose to share the forward-KL information with the discriminator by explicitly sharing the parameters between the encoder and the discriminator network. We restrict the sharing of the parameters between the encoder and discriminator networks until the penultimate layer to facilitate the modelling of the different outputs. 

Figure \ref{fig:approach_onecol} illustrates our proposed two-tier architecture and contrasts it against the traditional hybrid (V)AE/GAN approaches. 
We propose an adversary free two-tier architecture \ie an encoder and decoder network that has the capabilities of both a discriminator and a generator. 
In the proposed model the decoder network collapses into the generator, while the encoder network is merged with the discriminator resulting in a two tier architecture. We term our two tier architecture as Implicit Discriminator in Variational Autoencoder (IDVAE). VAE-GAN is a special case of IDVAE where there is no sharing of parameters between the encoder and the discriminator.

In this work we show that our proposed simpler hybrid VAE/GAN model, IDVAE outperforms the prior approaches in terms of visual fidelity measured in terms of FID score. We also show that our model sustains both reconstruction and generation ability without training an unnecessary adversary network that would result in learning more parameters. Overall the main contributions of the work are as follows : 
\begin{itemize}
    \item We introduce a novel two-tier architecture, IDVAE, which sustains the abilities of both reconstruction (VAE) and generation (GAN) without learning a separate discriminator or a generator.
    \item We present a training schedule that facilitates the encoder to act as as implicit discriminator while maintaining the tight coupling between encoder and decoder network.
    \item Empirical evaluations of IDVAE performed on benchmark datasets show that IDVAE achieves better generative ability than prior approaches. We also show that Fr\'echet Inception Distance, a common measure to evaluate the quality of the generations has inconsistent outputs and thereby propose an ensemble of experts for conducting quantitative evaluation.
\end{itemize}





\section{Related Work}

Variational Autoencoder (VAE) introduced by Kingma et al.\cite{VAE_kingma} 
minimizes the KL divergence between the real distribution ($P_x$) and the generated distribution ($P_g$) through the variational bound. 
Detailed analysis of VAE by Doersch\cite{vae_tutorial_doersch2016tutorial} shows that VAE works well in practice and is considered to model the true data distribution quite well but often generates poor quality samples \ie the images produced by the decoder are blurred. 
On the other hand GAN's \cite{Goodfellow:2014:GAN:2969033.2969125} generate samples that are visually more realistic through an adversarial game play between the generator and the discriminator. However, GAN's suffer from problems like instability during training and mode collapse.
Recently Bang and Shim \cite{DBLP:improved_training_using_representative_features} proposed RFGAN that uses pre-trained encoder features (representative features) to regularize the training of the discriminator to alleviate the problem of mode collapse. Similarly, MR-GAN\cite{mrgan_che2016mode} also proposes to use autoencoder features as regularizer in GAN training. Inspired from these architectures, we propose IDVAE that exploits the complementary properties of forward KL and reverse KL to capture the data distribution. While these approaches make use of a pre-trained encoder, our approach jointly and simultaneously trains a VAE and GAN achieving both the reconstruction and generation capabilities.  
There has also been some efforts towards utilizing the advantages of VAE for training GANs. Larsen \etal \cite{vae_gan_DBLP:journals/corr/LarsenSW15} proposed VAE-GAN that collapses the decoder of the VAE into the generator of the GAN. VAE-GAN achieves sharp generations using a similarity metric learned by the intermediate representations of an explicit adversary. VAE-GAN requires an explicit discriminator, while our proposed approach overcomes this necessity by converting the encoder of the network into a discriminator. ALI\cite{ALI_dumoulin2017adversarially-iclr} and BiGAN\cite{BIGAN_DBLP:journals/corr/DonahueKD16} also propose to use three networks: the encoder, the decoder and the adversary. Unlike IDVAE, both the ALI and the BiGAN discriminator differentiates between samples from the joint distribution of observed data and latent codes. However, the reported reconstructions are of poor quality \cite{Mescheder2017ICML}. Akin to BiGAN discriminator setting, the AVB\cite{Mescheder2017ICML} model uses an additional discriminator to facilitate learning without explicitly assuming any form for posterior distribution. 
However, the samples generated by AVB for the CelebA dataset are observed to be blurry 
\cite{AVB_github_repo}. In contrast, the simpler IDVAE model is able generate higher quality samples with lesser parameters.

Li \etal \cite{ALICE_li2017alice} propose ALICE that improves upon ALI by alleviating certain undesirable solutions (saddle points). Unlike a two tier approach of IDVAE, ALICE requires three networks and proposes to regularize the objective with cycle loss (an upper bound for conditional entropy). While in IDVAE, there is an implicit regularization on the discriminator by sharing its parameters with the encoder. The AS-VAE model of Pu \etal \cite{ASVAE_NIPS2017_7020} focuses on both reverse and forward-KL between the encoder and decoder joint distributions with an objective to maximize the marginal likelihood of observations and latent codes. AS-VAE also needs two adversaries to circumvent the need of assuming an explicit form for the true intractable distribution (eqn 8 and 9 in \cite{Mescheder2017ICML}). IDVAE also focuses on forward-KL and reverse-KL but in a very novel way by sharing the parameters of the encoder (forward-KL) and discriminator (reverse-KL) resulting in a simpler model. 
$\alpha$-GAN \cite{rosca2017variational} fuses VAE and GAN exploiting the density ratio trick by constructing two additional discriminators for measuring the divergence between the reconstructions and the true data points, and the latent representations and the latent prior. The first discriminator minimizes the reverse-KL divergence, and the reconstruction error term minimizes the forward-KL divergence to discourage mode collapse. Training $\alpha$-GAN is difficult as it requires learning a large set of parameters (for the 4 networks). 
In contrast to previous approaches, Ulyanov \etal \cite{AGE_DBLP:conf/aaai/UlyanovVL18} propose a two tier adversary free approach, AGE, where the encoder network is responsible for the adversarial signal. While the architectures of AGE and IDVAE appear to be similar, there are some fundamental differences in the process of learning the discriminator. The AGE discriminator compares (via divergence) the encoded real and fake distributions against a fixed reference distribution (typically, a prior in latent space). Whereas the IDVAE discriminator directly compares the real and fake data using a simple cross entropy loss, where both the reconstructions and randomly generated samples are treated as fake examples. We empirically show that IDVAE learns better as its discriminator relies on reconstructed samples as well. Importance of reconstructed samples in adversarial learning is supported in literature\cite{MRGAN_DBLP:journals/corr/CheLJBL16}.

\section{Methodology}

\begin{figure}[t]
\begin{center}
   \includegraphics[width=\linewidth]{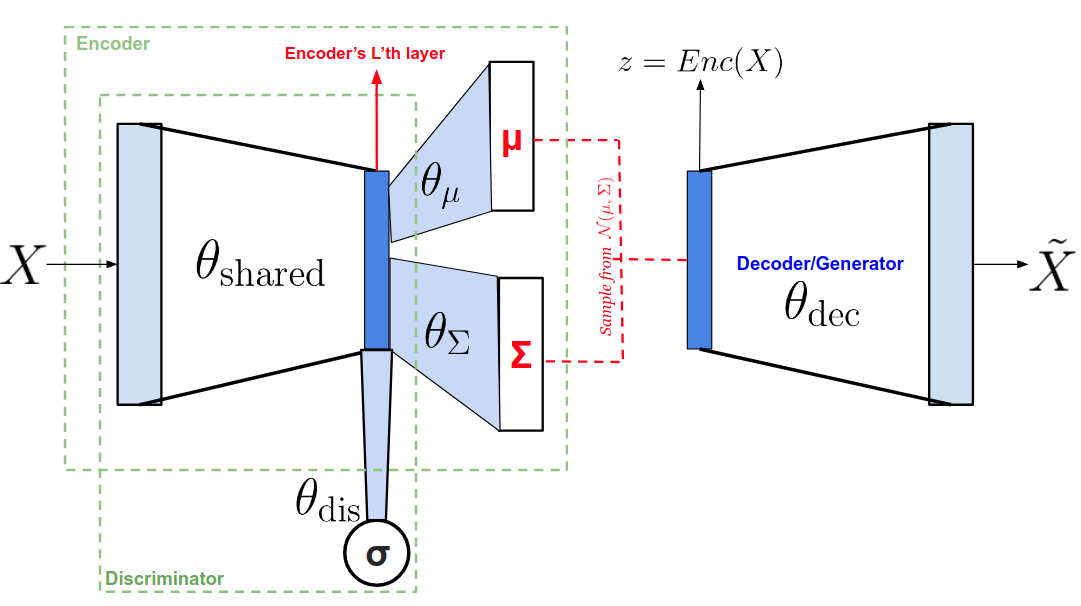}
\end{center}
   \caption{The proposed architecture for IDVAE. The parameters $\theta_{\mu}$ and $\theta_{\Sigma}$ denote a single fully connected layer learning the encoder specific parameters, $\theta_{\text{enc}}$. $\theta_{\text{dis}}$ also represents a single fully connected layer which denotes discriminator specific parameters. Similarly $\theta_{\text{shared}}$ denotes the shared parameters between the encoder and the discriminator whereas $\theta_{dec}$ denotes the decoder/generator specific parameters.}
\label{fig:long}
\label{fig:onecol}
\end{figure}

\textbf{Notations} Let $\textbf{x}$ be the data point in the input space $\mathcal{X}$ and $\textbf{z}$ be the code in the latent space $\mathcal{Z}$. The output of the encoder and the discriminator network for an input $\textbf{x}$ is represented as $\text{Enc}(\textbf{x})$ and $\text{Dis}(\textbf{x})$ respectively. Similarly the output of decoder network \ie $\Tilde{\textbf{x}}$ for a latent code $\textbf{z}$ is denoted by $\text{Dec}(\textbf{z})$. The output at the $l^{\text{th}}$ layer of the encoder network for an input $\textbf{x}$ is denoted as $\text{Enc}_{l}(\textbf{x})$. This is same as the output at the $l^{\text{th}}$ layer of the discriminator network for an input $\textbf{x}$ which is denoted as $\text{Dis}_{l}(\textbf{x})$. $\text{Enc}_{l}(\textbf{x})$ and $\text{Dis}_{l}(\textbf{x})$ are used interchangeably depending on the context.
In reference to Figure \ref{fig:onecol}, we denote the encoder specific parameters by $\theta_{enc}$ where $\theta_{enc} = \{\theta_{\mu},\theta_{\Sigma}\}$.

We start with some preliminaries on VAE\cite{VAE_kingma} and  GAN\cite{Goodfellow:2014:GAN:2969033.2969125} before describing our proposed model, which combines both of them. 
\subsection{Variational Autoencoder}
A VAE comprises of learning two networks, namely, the encoder and the decoder network. In contrast to traditional autoencoders, VAE views the encoder and the decoder networks as probabilistic functions. The encoder learns a conditional distribution on the latent code $\textbf{z}$ conditioned on the input $\textbf{x}$. Similarly decoder learns a distribution on $\Tilde{\textbf{x}}$  conditioned on the latent code $\textbf{z}$.

\begin{equation}
    \textbf{z} \thicksim \text{Enc}(\textbf{x}) = q(\textbf{z}|\textbf{x})
\end{equation}
\begin{equation}
    \Tilde{\textbf{x}} \thicksim \text{Dec}(\textbf{z}) = p(\textbf{x}|\textbf{z})    
\end{equation}

Vanilla VAE jointly trains over the encoder and the decoder network parameters by minimizing negative log-likelihood (reconstruction term) and divergence between prior and learned distribution in latent space $\mathcal{Z}$. The prior, $p(\textbf{z})$, over the latent space is typically assumed to be a unit Normal distribution, i.e. $\textbf{z}\thicksim \mathcal{N}(\textbf{0}, \textbf{I})$. Thus training a VAE would mean minimizing the following loss:
\begin{equation}
    \mathcal{L}_{\text{VAE}} = \mathcal{L}_{\text{recons}} + \mathcal{L}_{\text{prior}}
\end{equation}
where,
\begin{equation}\label{eqn:eq_4}
    \mathcal{L}_{\text{recons}} = -\mathds{E}_{q(\textbf{z}|\textbf{x})}[\log p(\textbf{x}|\textbf{z})]
\end{equation}
\begin{equation}\label{eqn:eq_5}
    \mathcal{L}_{\text{prior}} = KL(q(\textbf{z}|\textbf{x})\|p(\textbf{z}))
\end{equation}
and $KL(q(\textbf{z}|\textbf{x})\|p(\textbf{z}))$ is the Kullback-Leibler divergence between the distributions $q(\textbf{z}|\textbf{x})$ and $p(\textbf{z})$.
\subsection{Generative Adversarial Network}

A GAN consists of two networks, namely, the discriminator and the generator that are learned through an adversarial game play.
The generator network maps a point $\textbf{z}$ in some arbitrarily low dimensional latent space $\mathcal{Z}$ to a point in a high dimensional data space $\mathcal{X}$. We denote $\text{Gen}(\textbf{z})$ as the output of the generator network when $\textbf{z}$ is the input. In a similar vein, the discriminator network maps a data point $\textbf{x}$ in the data space to a probability value $y \in [0,1]$. The objective of the discriminator is to assign the probability $y=\text{Dis}(\textbf{x})$ that $\textbf{x}$ is a sample from true distribution 
and the probability $1-y$ that $\textbf{x}$ is a generated sample i.e. $\textbf{x}=\text{Gen}(\textbf{z})$, with $\textbf{z}\thicksim p(\textbf{z})$. Thus, in this adversarial game play, the objective of the generator is to synthesize samples that can fool the discriminator \ie learning the true data distribution, while the goal of the discriminator is to recognize the samples coming out of the generated(fake) distribution and the true distribution. Adversarial game play between the discriminator and the generator is formally defined by the GAN loss as
\begin{equation}\label{eqn:eq_6}
     \mathcal{L}_{\textit{GAN}} = \log(\text{Dis}(\textbf{x})) + \log(1-\text{Dis}(\text{Gen}(\textbf{z})))
\end{equation}
 We want to maximize the binary cross entropy loss with respect to the discriminator (D) while minimizing it for the generator (G). Thus, the minimax objective is defined as
\begin{equation}\label{eqn:eq_6_2}
     \underset{G}{\text{min}}\, \underset{D}{\text{max}}\,\mathcal{L}_{\textit{GAN}}
\end{equation}

\subsection{IDVAE}

Our proposed approach, \textbf{I}mplicit \textbf{D}iscriminator in \textbf{V}ariational \textbf{A}uto\textbf{E}ncoder (IDVAE), exploits the properties of both VAE and GAN. IDVAE sustains the stable training properties of VAE while generating samples of quality approaching GAN. 
We borrow the encoder and decoder networks from VAE with slight modifications. In particular, we collapse the VAE decoder network into the generator of the GAN and the VAE encoder network partially into the discriminator of the GAN.

We partially collapse the encoder into the discriminator following the assumption that there exists an overlap in the knowledge of encoder and discriminator network. As the encoder's objective is to learn representational features, while the discriminator's objective is to learn discriminative features, we restrict the weight sharing to the penultimate layer (say some $l^{th}$ layer of encoder represented as $\text{Enc}_{l}$) in encoder of the VAE. Further to encourage the encoder to learn the features of discriminator we add a single fully connected layer from $\text{Enc}_l$ to a single sigmoid node that acts as the discriminator's output. Figure \ref{fig:onecol} illustrates the proposed IDVAE network architecture.

Thus, in our model we have four sets of parameters that need to be learned, namely; $\theta_{\text{dec}}$ - the shared parameters between the decoder and the generator, $\theta_{\text{shared}}$ - the parameters shared between the encoder and the discriminator, $\theta_{\text{enc}}$ - the encoder specific parameters of the VAE, and $\theta_{\text{dis}}$ - the discriminator specific parameters of the GAN. These are updated based on the loss incurred by each of the individual networks. 

The loss incurred by the encoder is used to update both $\theta_\text{shared}$ and $\theta_\text{enc}$. The encoder loss in the IDVAE consists of two components similar to a standard VAE. The first component is the reconstruction loss - $\mathcal{L}_{\text{recons}}$ and the second component is the prior discrepancy loss - $\mathcal{L}_{\text{prior}}$. It is well known 
that minimizing the forward-KL divergence \ie KL($P_{data}\|P_{model}$) achieves mode coverage for generative models. Thus, we minimize forward-KL divergence by minimizing $\mathcal{L}_{\text{recons}}$ for helping IDVAE to learn the different modes in the data. 
The shared parameters between the encoder and the discriminator encode the forward-KL divergence information. Thus using $\mathcal{L}_{\text{recons}}$ in the encoder reduces the extent of mode collapse as the gradients from the discriminator to the generator implicitly contain the information about multiple modes. 
Thus the overall encoder loss ($\mathcal{L}_{enc}$) is defined as follows
\begin{equation}\label{eqn:eq_9}
    \mathcal{L}_{\text{enc}} = \alpha\mathcal{L}_{\text{recons}} + \beta\mathcal{L}_{\text{prior}}
\end{equation}

where $\alpha$ and $\beta$ are hyper parameters 
controlling the contribution of each of the loss terms.

It has been shown that GAN \cite{Goodfellow:2014:GAN:2969033.2969125} achieves sharper images by minimizing the reverse-KL divergence. Thus, we use the implicit adversary (encoder as a shared discriminator) of IDVAE as a way to propagate reverse-KL divergence information. The discriminator loss is used to update both $\theta_\text{shared}$ and $\theta_\text{dis}$. The generated (fake) examples that are presented to the discriminator of IDVAE are the output of the decoder when viewed as a generator $\text{Dec}(\textbf{z})$, where $\textbf{z}\thicksim p(\textbf{z})$. In addition to this, we also present the synthesized sample through reconstruction, $\text{Dec}(\text{Enc}({\textbf{x})})$, for an input $\textbf{x}$. As $\text{Dec}(\text{Enc}(\textbf{x}))$ is more likely to be similar to $\textbf{x}$ than $\text{Dec}(\textbf{z})$, for an arbitrary $\textbf{z} \thicksim p(\textbf{z})$, we hypothesize that the discriminator loss corresponding to $\text{Dec}(\text{Enc}(\textbf{x}))$ encourages the generator to learn the properties of the decoder. Similarly the discriminator loss corresponding to $\text{Dec}(\textbf{z})$ encourages the decoder to learn the properties of the generator \ie be able to generate realistic examples from the prior distribution $p(\textbf{z})$. Therefore using both the terms, $\text{Dec}(\textbf{z})$ and $\text{Dec}(\text{Enc}({\textbf{x})})$ encourages the model to learn a blend of both the generator and the decoder. Intuitively in equation \ref{eqn:eq_10} to maintain the ratio of real and fake samples shown to the discriminator the loss terms for the fake samples should be scaled by a factor of 0.5 or the real term \ie $\text{log(Dis(\textbf{x}))}$ by 2. We observed no significant change in the performance of IDVAE when these factors are dropped, thereby giving rise to the following loss function
\begin{multline}\label{eqn:eq_10}
     \mathcal{L}_{\text{dis}} = -\big[\log(\text{Dis}(\textbf{x})) +\log(1-\text{Dis}(\text{Dec}(\textbf{z}))) \\
     +\log(1-\text{Dis(Dec(Enc($\textbf{x}$))})\big]
 \end{multline}

As we have collapsed the decoder of the vanilla VAE into the generator, the loss incurred by both the decoder and generator is used to update the shared parameters between the decoder and generator ($\theta_\text{dec}$). The decoder/generator loss in IDVAE consists of two components. 
The first component ($\mathcal{L}^\text{dis}_\text{recons}$) is a learned similarity metric motivated by VAE-GAN \cite{vae_gan_DBLP:journals/corr/LarsenSW15}. Specifically, we learn a similarity metric($\mathcal{L}_{\text{recons}}^{\text{dis}}$) using an intermediate representation ( $l^{\text{th}}$ layer) of the discriminator (equivalent to the $l^{\text{th}}$ layer of the encoder) by assuming a Gaussian observation model on $\text{Dis}_{l}(\Tilde{\textbf{x}})$ with mean $\text{Dis}_{l}(\textbf{x})$ and unit covariance : 
 \begin{equation}\label{eqn:eq_7}
     p(\text{Dis}_{l}(\Tilde{\textbf{x}})|\textbf{z}) = \mathcal{N}(\text{Dis}_{l}(\Tilde{\textbf{x}})|\text{Dis}_{l}(\textbf{x}),\textbf{I})
 \end{equation}
where for a given sample \textbf{x}, $\Tilde{\textbf{x}} = \text{Dec}(\textbf{z})$ and $\textbf{z} = \text{Enc}(\textbf{x})$. 
$\mathcal{L}_{\text{recons}}^{\text{dis}}$ is defined as a Gaussian observation model: 
\begin{equation}\label{eqn:eq_8}
    \mathcal{L}_{\text{recons}}^{\text{dis}} = -\mathds{E}_{q(\textbf{z}|\textbf{x})}[\text{log }p(\text{Dis}_{l}(\textbf{x})|\textbf{z})]
\end{equation}
The second component is the adversarial loss which encourages the decoder to learn the properties of a generator. The adversarial loss, ($\mathcal{L}_{\text{GAN}}$), is defined as
\begin{equation}\label{eqn:eq_11}
    \mathcal{L}_{\text{GAN}} = -log(\text{Dis(Dec($\textbf{z}'$)}))- log(\text{Dis(Dec(Enc(\textbf{$\textbf{x}$})))})  
\end{equation}
where $\textbf{z}' \thicksim p(\textbf{z})$.

Therefore the presence of both the reconstruction loss and the adversarial loss in objective function of decoder makes it learn a blend of the two models. The overall loss function for the decoder/generator ($\mathcal{L}_\text{dec}$) is defined as:
\begin{equation}\label{eqn:eq_12}
     \mathcal{L}_\text{dec} = \omega\mathcal{L}_{\text{GAN}}
     +\lambda\mathcal{L}_{\text{recons}}^{\text{dis}}
\end{equation}

where $\omega \text{ and }\lambda$ used in $\mathcal{L}_\text{dec}$ are hyper-parameters that are learned empirically. 

\subsection{Training Schedule}
The $l^{th}$ shared layer between the encoder and the discriminator outputs representations for learning the parameters of the encoder's distribution and for discriminating between samples from the true distribution and generated samples simultaneously. We need to ensure that the shared weights ($\theta_{\text{shared}}$) of the encoder and discriminator network gets the learning signal corresponding to both the encoder and discriminator objective function. In theory $\theta_{\text{shared}},\theta_{\text{dis}}$, and $\theta_{\text{enc}}$ can be updated in a single step using the joint loss of both the encoder and discriminator. However, in practise we observed training using the joint loss to be challenging in terms of hyper-parameter fine tuning. Hence, we update the shared weights ($\theta_{\text{shared}}$) in two iterations, once each with the encoder and discriminator losses respectively. In the first iteration $\theta_{\text{shared}}$ and $\theta_\text{dis}$ are updated while in the second iteration $\theta_\text{shared}$ and $\theta_\text{enc}$ are updated. Algorithm~\ref{alg:train_schedule} presents the overview of the training procedure. Thus the parameters $\theta_\text{shared}$ learn the information of both reverse-KL (first iteration) and forward-KL (second iteration), which is leveraged by the decoder/generator in the third step of the algorithm. The parameters,  $\theta_\text{shared}$, can be updated in any arbitrary order, we empirically found that using first the discriminator loss helps in better learning. We present the qualitative results on the other two variants (using the joint loss, and updating $\theta_{\text{shared}}$ with respect to the encoder first followed by the discriminator) in the supplementary material.



\begin{algorithm}[t]
	\caption{IDVAE Training Schedule}
	\label{alg:train_schedule}
	\begin{algorithmic}[1]
        \STATE $P(z)\leftarrow \mathcal{N}(0,\textit{I}) $\\
       \STATE $\theta_\text{shared},\theta_\text{enc},\theta_\text{dec},\theta_\text{dis}\leftarrow \text{Initialize parameters}$
        \STATE $X \gets \text{random mini batch from dataset} $
        \STATE $Z \gets \text{Enc($X$)} $
        \STATE $\Tilde{X} \gets \text{Dec($Z$)} $
        \STATE $Z' \gets \text{samples from prior }P(Z) $
        \STATE $X' \gets \text{Dec($Z'$)} $
       
        \WHILE {\textit{not convergence}}
            \STATE $\theta_\text{dis},\theta_\text{shared} \xleftarrow[]{+} -\nabla_{(\theta_\text{dis},\theta_\text{shared})}(\mathcal{L}_\text{dis})$
            
            \STATE $\theta_\text{enc},\theta_\text{shared} \xleftarrow[]{+} -\nabla_{(\theta_\text{enc},\theta_\text{shared})}(\mathcal{L}_\text{enc})$
            
            \STATE 
            $\theta_\text{dec} \xleftarrow[]{+} -\nabla_{\theta_\text{dec}}(\mathcal{L}_\text{dec})$
            
        \ENDWHILE
	\end{algorithmic}
\end{algorithm}

\section{Experiments}
We investigate the proposed IDVAE architecture for the quality of both reconstructions and generations. 
We evaluate the performance of IDVAE on the following two real world benchmark datasets: 
i) \textbf{CIFAR10} \cite{cifar10_krizhevsky2009learning}, which contains 60k images of which 50k are used for training and the remaining 10k for testing. 
(ii) \textbf{CelebA} \cite{celebA_liu2015faceattributes}, which consists of 202,599 images. We use 1-162,770 images for training, 162,771-182,637 for validation and rest for testing. In our implementation pipeline we crop and scale the images to 64x64 for faster training. The details of encoder and decoder network architecture along with fine tuned hyper-parameters for each of the dataset are provided in the supplementary material. 
For generating instances over the different datasets we randomly sample $\textbf{z}$ from the assumed prior distribution (on the latent space $\mathcal{Z}$) $\mathcal{N}(\textbf{0},\textbf{I})$. We also conducted experiments on a synthetic 2D Gaussian dataset and the MNIST digits dataset. These details can be found in the supplementary material.

We compare the performance of IDVAE against approaches that have both generative and reconstruction abilities, namely;  VAE\cite{VAE_kingma}, VAE-GAN\cite{vae_gan_DBLP:journals/corr/LarsenSW15}, AGE \cite{DBLP:journals/corr/UlyanovVL17a}, and $\alpha$-GAN\cite{rosca2017variational}. We use the pre-trained models available for AGE, while we train all the other models from scratch using the best hyper-parameters reported in the literature.

\section{Results and Discussion}
We conduct both qualitative and quantitative comparison of IDVAE for generations and reconstructions against all the prior approaches.

\subsection{Quantitative Analysis}
\begin{figure}[t]
    \centering
    \includegraphics[width=0.5\textwidth]{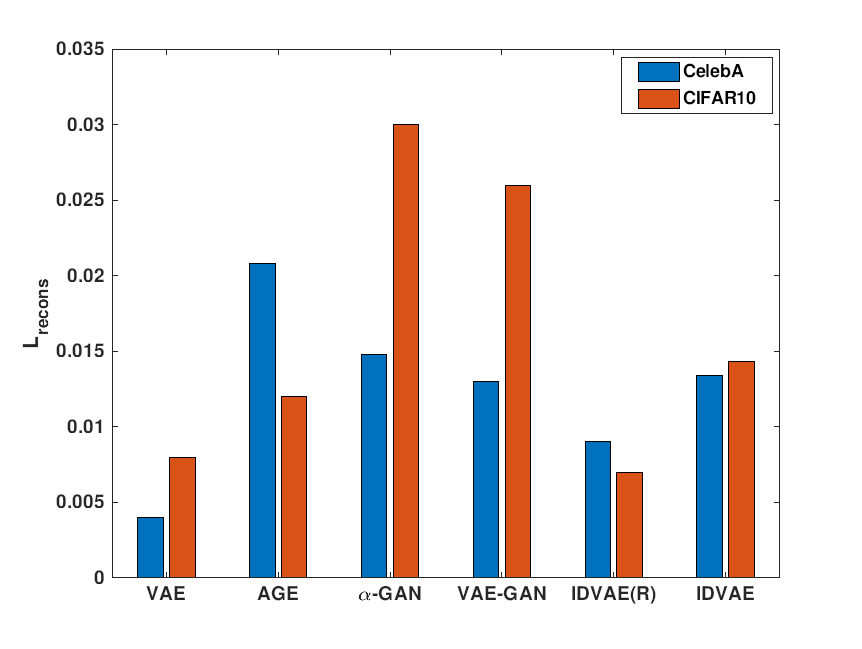}
    \caption{Comparing reconstruction loss (lower is better) among different generative models.}
    \label{fig:recons_scores}
\end{figure}

The reconstruction quality is objectively quantified using the standard square loss $\mathcal{L}_\text{recons}$. We obtain an unbiased estimate of the loss using a large test set consisting of 10k samples for both the CelebA and the CIFAR10 datasets. Thus even a small improvement on such a large set is significant considering the complexities of the dataset. 
It is evident from the results presented in Figure \ref{fig:recons_scores} that for the reconstruction task IDVAE performs better than or is at par with VAE-GAN, AGE and $\alpha$-GAN. However the lowest reconstruction error is obtained by VAE. This is understandable as there is no explicit penalty on the decoder for reconstructing unrealistic images. On the other hand both VAE-GAN and IDVAE strike a balance between the reconstruction loss and the generative ability. We also modify IDVAE to explicitly minimize $\mathcal{L}_{\text{recons}}$ in the decoder, which we term as IDVAE(R). The reconstruction loss of IDVAE(R) is the best among all the variants of VAE, However, this comes at the cost of quality of the generations.

There are two popular measures for qualitatively evaluating generative models, namely the Inception Score (IS) \cite{inception_score_salimans2016improved} and Fr\'echet Inception Distance (FID) \cite{DBLP:FID_journals/corr/HeuselRUNKH17}. It has been shown that IS closely follows human scoring of images synthesized by generative models for the CIFAR 10 dataset\cite{cifar10_krizhevsky2009learning}. The IS uses the Inception v3 model pre-trained on ImageNet. The IS is a statistic on the Inception model's output when applied to the synthesized images. This statistic captures two desirable qualities of a generative model - the synthesized images should contain an object (the image is sharp and not blurry) that is reflected in a low entropy output of the Inception model\cite{inception_model_szegedy2015going} and; there must be diversity in the generations that is reflected in the high entropy output of the Inception model over the entire generated set. Barratt and Sharma \cite{note_on_IS_barratt2018note} have recently shown that the IS suffers from suboptimalities and is an appropriate measure only for datasets that are trained on ImageNet. Thus, it is not advisable to measure the quality of generations on the CelebA dataset. The FID improves upon the IS by comparing the statistics of both the generated and true samples, instead of evaluating the generated samples in isolation. The FID is the Fr\'echet distance between two multivariate Gaussians estimated from the 2048-dimensional activations of the Inception-v3 pool3 layer for real and generated samples. Lower FID scores correspond to more similar real and synthetic samples.

 Table \ref{tab:FID_scores}, presents the Fr\'echet distance scores computed using Inception-15\footnote{weights used from 2015 year model \cite{inception2k15}} ($FID_{15}$), Inception-16\footnote{weights used from 2016 year model \cite{inception2k16}} ($FID_{16}$), and ResNet\cite{resnet_he2016deep}. As all the three experts have the same knowledge \ie all the models are pre-trained on ImageNet\cite{imagenet_cvpr09} and the representations extracted from the intermediate layer have the same dimensions (2048), the relative performance of the models with respect to each expert can be compared. While the distance scores across the experts for the same model may be different, we expect the order of the goodness among the generative models to be preserved. However, as can be observed from Table \ref{tab:FID_scores} on CIFAR10 dataset, VAE-GAN appears to perform better than the AGE based on $FID_{15}$, while the trend reverses when comparing based on both $FID_{16}$ and $FRD$. Therefore, 
 our results suggest that a generative model should be compared across a battery of experts rather than in isolation. 
 IDVAE performs better than all the other approaches on the CelebA dataset. The result is statistically significant on both $FID_{15}$ and $FRD$ scores. On the other hand both IDVAE and $\alpha$-GAN result in the best performance on the CIFAR10 dataset. There is no significant difference between IDVAE and $\alpha$-GAN with each performing better than the other only according to a single measure. However, from Figure \ref{fig:recons_scores}, it is quite apparent that $\alpha$-GAN focuses less on reconstructions whereas in IDVAE we do not observe such a bias. These results support our hypothesis that the encoder and discriminator can be a shared network. IDVAE is able to perform at par or sometimes better than VAE-GAN and $\alpha$-GAN that require a separate encoder/discriminator network. This is further verified through our qualitative results. We also observe a dip in the Fr\'echet distance for the IDVAE(R) model in comparison to IDVAE. As the decoder/generator of IDVAE(R) focuses on reconstructions in the image space we observe a drop in the Fr\'echet distance at the cost of a better reconstruction loss. Therefore IDVAE(R) model has the potential to fit within the required thresholds by tuning the hyper parameters $\omega, \lambda \text{ and } \gamma$.

\begin{table}[t]
\begin{center}
\small
\begin{tabular}{|l|c|c|c|}
\hline
Model & $\text{FID}_{15}$ & $\text{FID}_{16}$ & FRD \\
\hline \hline
\multicolumn{4}{|c|}{CIFAR10}\\
\hline
True Data & 3.16$\pm$0.06 & 7.41$\pm$0.82 & 26.17$\pm$1.44\\
IDVAE & 23.48$\pm$0.15 & 28.15$\pm$0.39 & \textbf{105.45$\pm$0.79}\\
IDVAE(R) & 43.38$\pm$0.15 & 49.9$\pm$0.85 & 191.32$\pm$5.88\\
VAE-GAN & 27.04$\pm$0.12 & 33.12$\pm$0.73 & 139.95$\pm$2.71\\
VAE & 85.74$\pm$0.3 & 130.38$\pm$3.47 & 626.67$\pm$8.61\\
AGE & 32.19$\pm$0.3 & 29.3$\pm$0.54 & 122.43$\pm$2.61 \\
$\alpha$-GAN & \textbf{20.61$\pm$0.12} & 27.87$\pm$0.7 & 121.88 $\pm$ 3.09\\
\hline
\multicolumn{4}{|c|}{CelebA}\\
\hline
True Data & 1.58$\pm$0.02 & 2.67$\pm$0.15 & 5.77$\pm$0.35 \\
IDVAE & \textbf{8.53$\pm$0.12} & 9.52$\pm$0.72 & \textbf{34.47$\pm$2.41} \\
IDVAE(R) & 14.81$\pm$0.17 & 16.71$\pm$0.66 & 70.99$\pm$0.91 \\
VAE-GAN & 9.52$\pm$0.06 & 10.5$\pm$0.9 & 38.32$\pm$1.69 \\
VAE & 35.27$\pm$0.04 & 55.44$\pm$0.87 & 150.02$\pm$1.41 \\
AGE & 12.74$\pm$0.14 & 15.27$\pm$0.36 & 82.45$\pm$0.97 \\
$\alpha$-GAN & 10.38$\pm$0.2 & 13.89$\pm$1.58 & 55.44$\pm$7.97\\


\hline
\end{tabular}\\[0.1cm]

\end{center}
 \caption{Comparing Frechet Distance (lower is better) among different generative models. 
 $FID_{15,16}$: Inception Model with 2015, 2016 year weights respectively and $FRD$ ResNet model.} 
\label{tab:FID_scores}
\end{table}

\subsection{Qualitative Analysis}
We present the qualitative results obtained from the different models in the Table \ref{tab:qualitative} on the CIFAR10 and CelebA datasets. It is quite evident from the images that VAE results in blurry reconstructions while the rest of the approaches output sharp images, which is due to presence of adversarial loss. On both CelebA and CIFAR10 datasets, we observe IDVAE and IDVAE(R) performing on par with VAE-GAN and $\alpha$-GAN, while significantly outperforming VAE in terms of sharpness of the images. The images generated by the different models are also presented in Table \ref{tab:qualitative}. The undesirable blurriness property in VAE is apparent on the CIFAR10 dataset while the performance of IDVAE is on par with both $\alpha$-GAN and VAE-GAN. The images generated by IDVAE trained on the CelebA dataset appears to capture a large diversity in background when compared to $\alpha$-GAN and VAE-GAN. We observe both IDVAE(R) and $\alpha$-GAN tend to focus more on faces than the background in these images. 
\begin{table*}[t]
    \begin{center}
    
  \begin{tabular}{ ccccc }
    Approach & \shortstack{\\CIFAR10\\ Reconstructions} & \shortstack{\\CIFAR10\\ Generations} & \shortstack{\\CelebA\\ Reconstructions} & \shortstack{\\CelebA\\ Generations} \\ 
    Original
    &
    \begin{minipage}{.18\linewidth}
      \includegraphics[width=\linewidth,height=3cm]{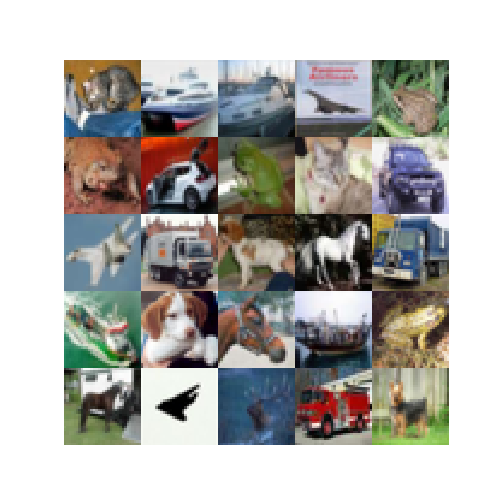}
    \end{minipage}
    &
    \begin{minipage}{.18\linewidth}
      \includegraphics[width=\linewidth,height=3cm]{images/cifar10/VAE/orig-img.png}
    \end{minipage}
    &
    \begin{minipage}{.18\linewidth}
      \includegraphics[width=\linewidth,height=3cm]{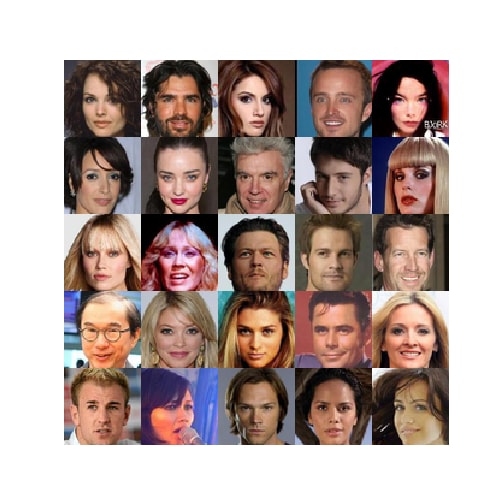}
    \end{minipage}
    &
    \begin{minipage}{.18\linewidth}
      \includegraphics[width=\linewidth,height=3cm]{images/celebA/VAE/orig_img.png}
    \end{minipage}
    \\ 
    VAE
    &
    \begin{minipage}{.18\linewidth}
      \includegraphics[width=\linewidth,height=3cm]{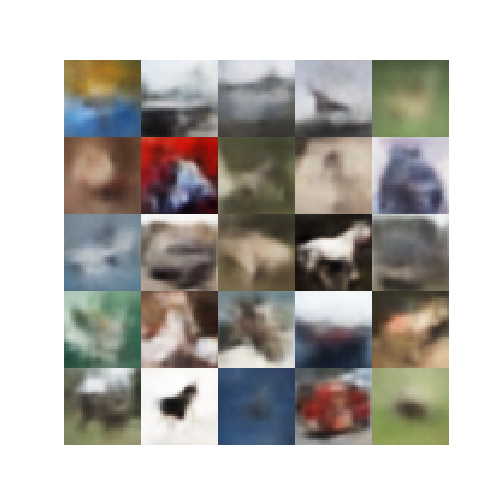}
    \end{minipage}
    &
    \begin{minipage}{.18\linewidth}
      \includegraphics[width=\linewidth,height=3cm]{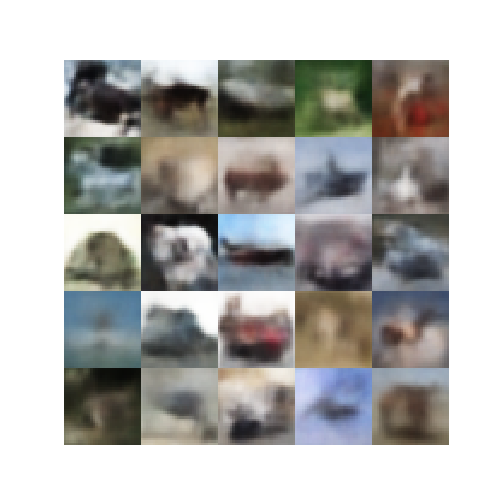}
    \end{minipage}
    &
    \begin{minipage}{.18\linewidth}
      \includegraphics[width=\linewidth,height=3cm]{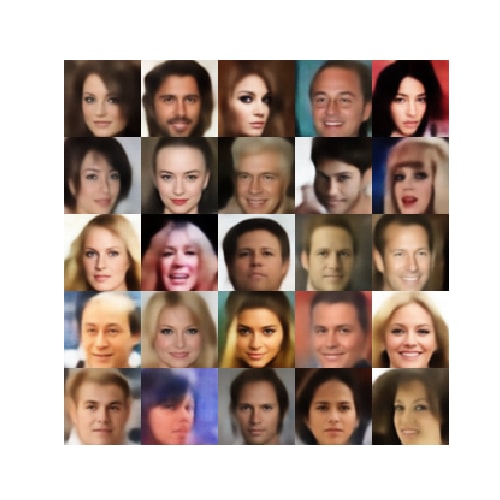}
    \end{minipage}
    &
    \begin{minipage}{.18\linewidth}
      \includegraphics[width=\linewidth,height=3cm]{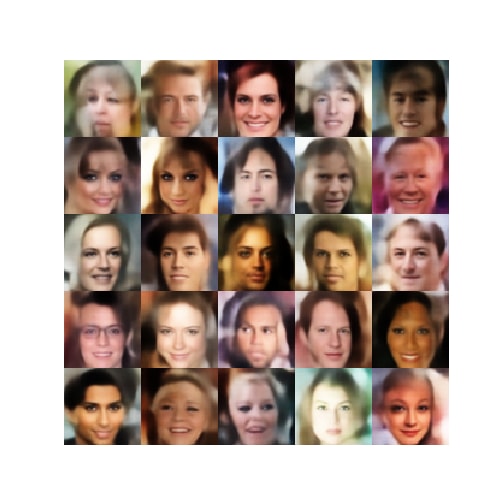}
    \end{minipage}
    \\ 
    IDVAE
    &
    \begin{minipage}{.18\linewidth}
      \includegraphics[width=\linewidth,height=3cm]{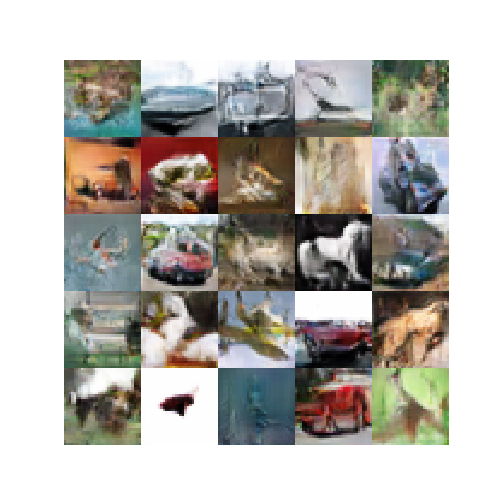}
    \end{minipage}
    &
    \begin{minipage}{.18\linewidth}
      \includegraphics[width=\linewidth,height=3cm]{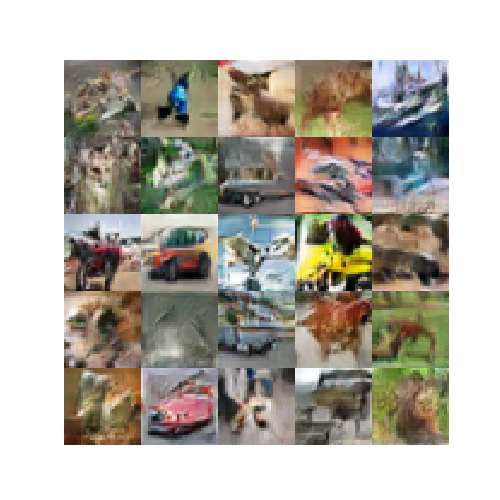}
    \end{minipage}
    &
    \begin{minipage}{.18\linewidth}
      \includegraphics[width=\linewidth,height=3cm]{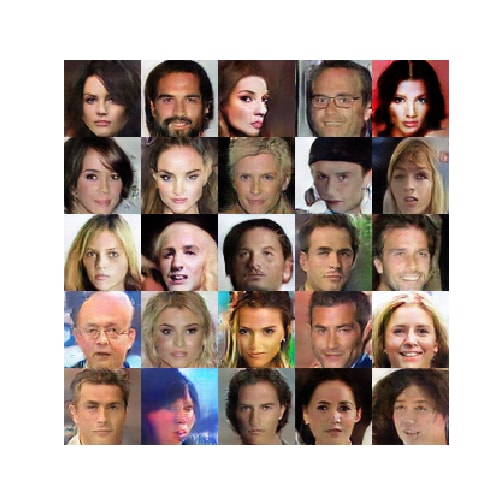}
    \end{minipage}
    &
    \begin{minipage}{.18\linewidth}
      \includegraphics[width=\linewidth,height=3cm]{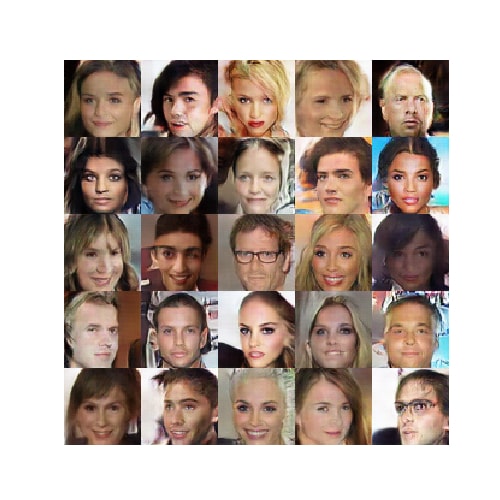}
    \end{minipage}
    \\ 
    
    IDVAE(R)
    &
    \begin{minipage}{.18\linewidth}
      \includegraphics[width=\linewidth,height=3cm]{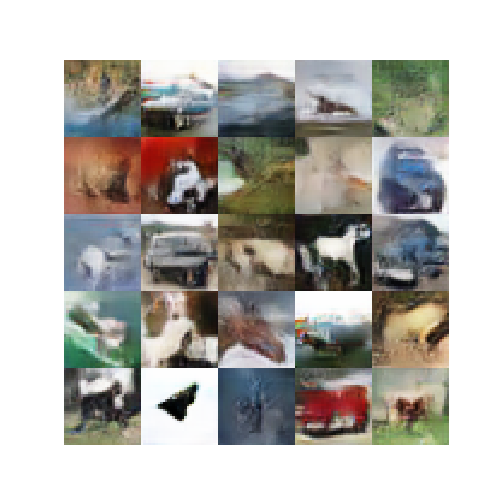}
    \end{minipage}
    &
    \begin{minipage}{.18\linewidth}
      \includegraphics[width=\linewidth,height=3cm]{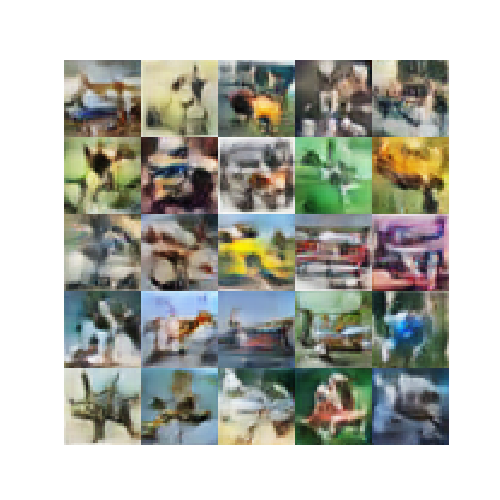}
    \end{minipage}
    &
    \begin{minipage}{.18\linewidth}
      \includegraphics[width=\linewidth,height=3cm]{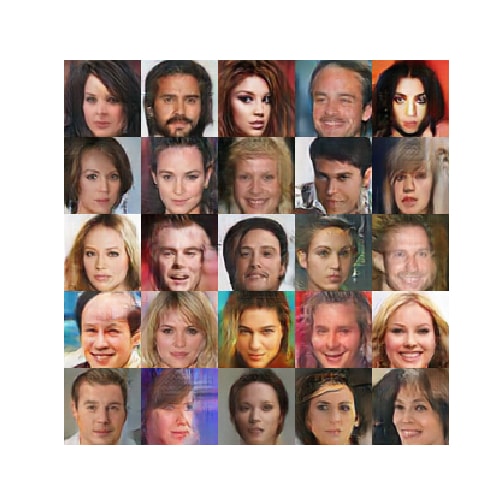}
    \end{minipage}
    &
    \begin{minipage}{.18\linewidth}
      \includegraphics[width=\linewidth,height=3cm]{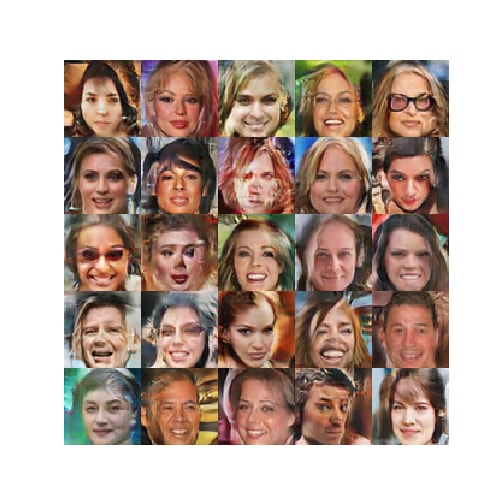}
    \end{minipage}
    \\ 
    
    VAE-GAN
    &
    \begin{minipage}{.18\linewidth}
      \includegraphics[width=\linewidth,height=3cm]{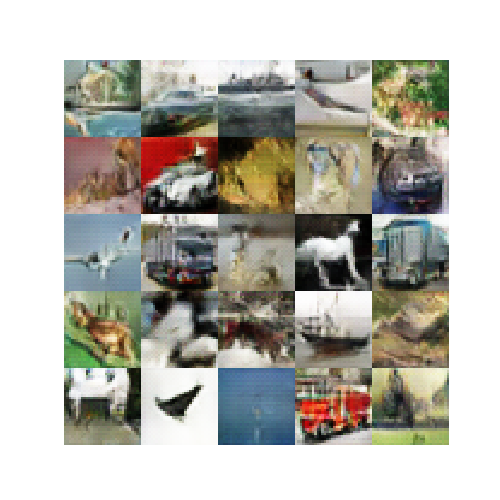}
    \end{minipage}
    &
    \begin{minipage}{.18\linewidth}
      \includegraphics[width=\linewidth,height=3cm]{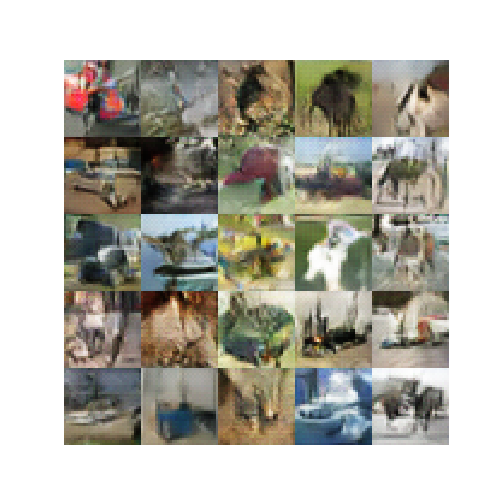}
    \end{minipage}
    &
    \begin{minipage}{.18\linewidth}
      \includegraphics[width=\linewidth,height=3cm]{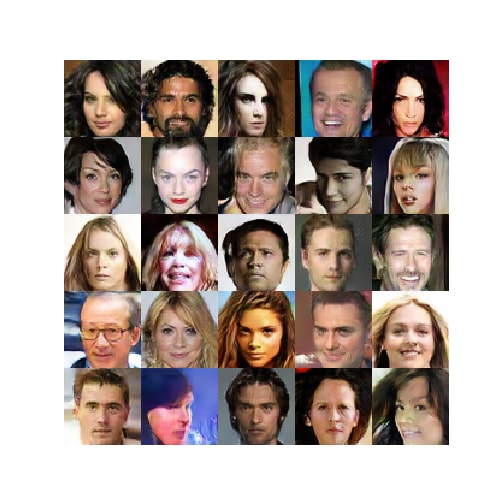}
    \end{minipage}
    &
    \begin{minipage}{.18\linewidth}
      \includegraphics[width=\linewidth,height=3cm]{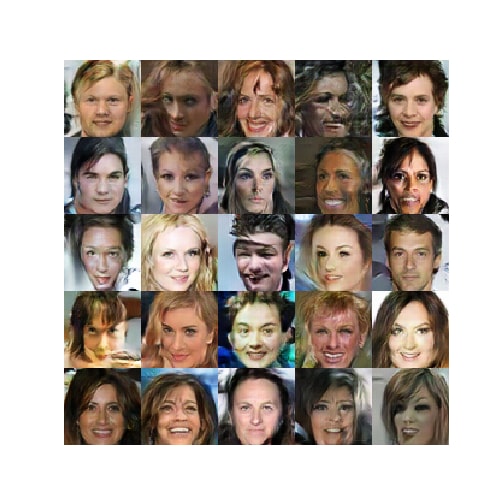}
    \end{minipage}
    \\ 
    
    $\alpha$-GAN
    &
    \begin{minipage}{.18\linewidth}
      \includegraphics[width=\linewidth,height=3cm]{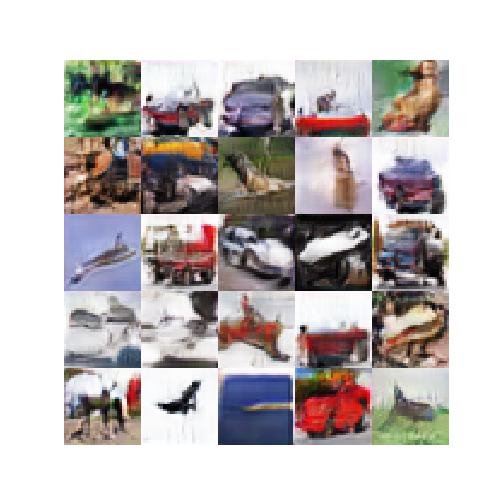}
      \end{minipage}
    &
    \begin{minipage}{.18\linewidth}
      \includegraphics[width=\linewidth,height=3cm]{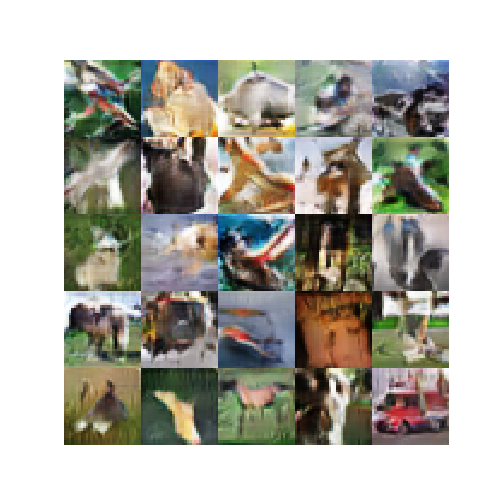}
      \end{minipage}
    &
    \begin{minipage}{.18\linewidth}
      \includegraphics[width=\linewidth,height=3cm]{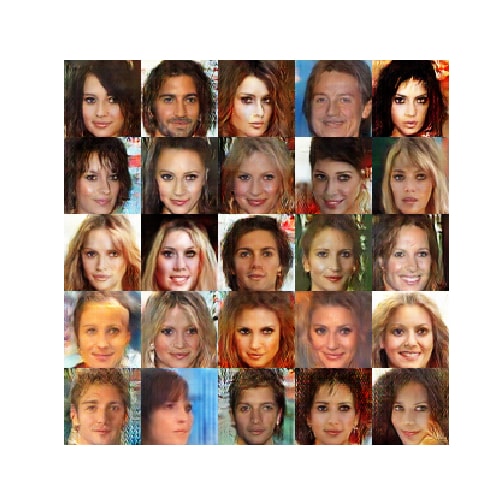}
      \end{minipage}
    &
    \begin{minipage}{.18\linewidth}
      \includegraphics[width=\linewidth,height=3cm]{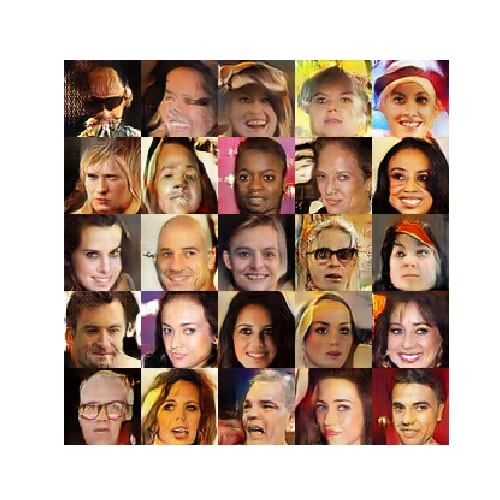}
      \end{minipage}
    \\ 
    
  \end{tabular}
  \caption{Qualitative experiments comparing different generative models.}
  \label{tab:qualitative}
  \end{center}
\end{table*}

\subsection{Conditional IDVAE}
We extend IDVAE to a conditional setting where our objective is to learn a generator/decoder whose output is controlled by some conditional information, $y$. We term this variant as \textbf{C}onditional-IDVAE (C-IDVAE). We qualitatively analyze the C-IDVAE model using MNIST\cite{mnist_LeCun1998GradientBasedLA} and CelebA datasets. We follow the recent work of Perarnau \etal \cite{icgan_perarnau2016invertible} to provide the conditional information to the generator at the input layer, while for the discriminator this is provided after the first convolution layer. 
To the best of our knowledge the encoder of VAE is never made aware of the conditional information but as we have our encoder acting as a discriminator we add this conditional information after the first convolution layer.

\begin{figure}[t!]
\begin{center}
  \includegraphics[width=0.65\linewidth]{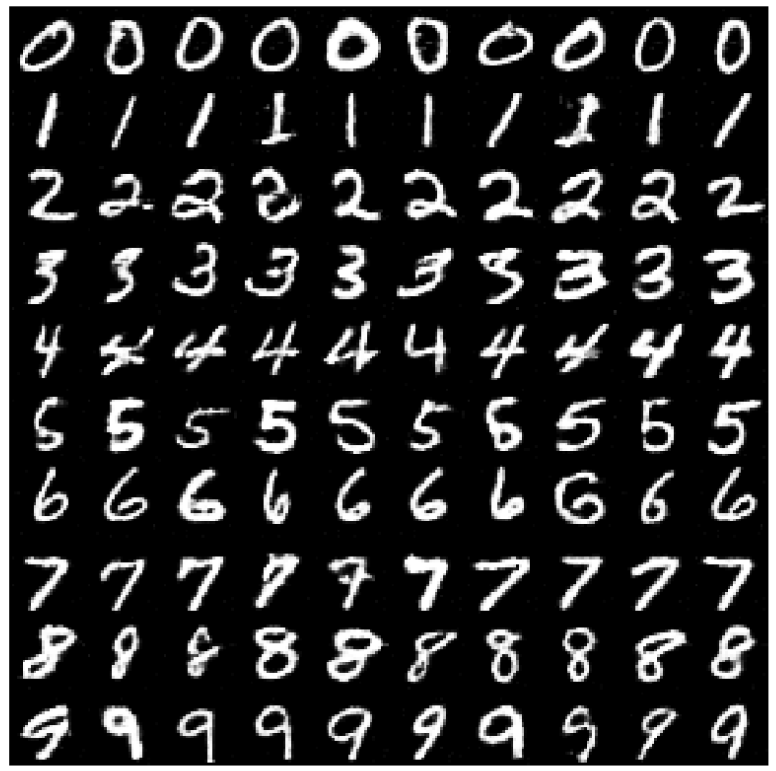}
  \includegraphics[width=0.65\linewidth]{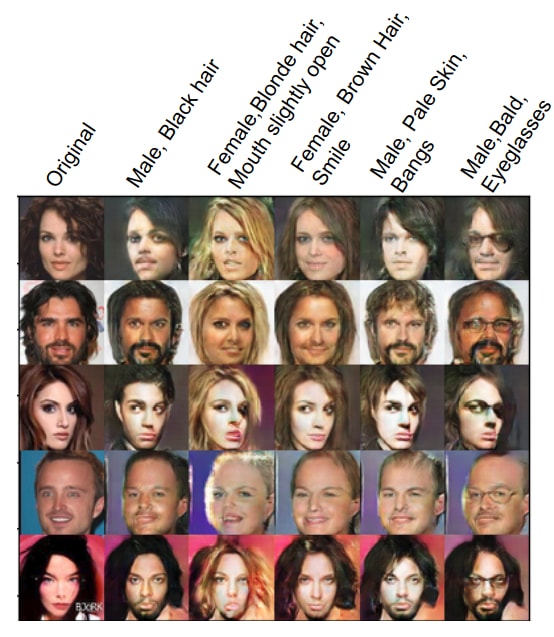}
  \caption{Conditional-IDVAE on MNIST followed by celebA conditioned on visual attributes.}
  \label{fig:exp_mnist_conditional_IDVAE}
\end{center}
\end{figure}

We use the one hot encoding of the MNIST class labels as the conditional information to evaluate C-IDVAE. Figure \ref{fig:exp_mnist_conditional_IDVAE} illustrates the samples generated by C-IDVAE, where each row illustrates the images generated by conditioning on a unique label. We observe a large diversity in the generations in each row implying the diverse generative ability of the conditional decoder/generator.

Following the work of Perarnau \etal \cite{icgan_perarnau2016invertible} we use 13 attributes that have clear visual impact out of the total 40 attributes as conditioning information while training C-IDVAE on the CelebA dataset. Figure \ref{fig:exp_mnist_conditional_IDVAE} presents the images generated by C-IDVAE for different conditioning information. Each row in the figure represents the images generated by C-IDVAE with the conditioning information provided in the top row, and the original image that is modified in the first column. It can be observed that the changes in each of the generations with respect to the original image are a result of the model imagining the original image on different attributes. For example, consider third and fifth rows in the Figure \ref{fig:exp_mnist_conditional_IDVAE}, where the original image is a female face. The generations in the columns 2, 5, and 6 that are conditioned on the male attribute actually contain a face that resembles a male. Similarly in the last column the model does reasonably well in adding eyeglasses to all the generated images. Thus considering the generative ability matching with the human imagination and the complexity of the real world CelebA dataset, C-IDVAE shows the potential to model the complex distributions.


\section{Summary and Future Work}
In this work we introduce a novel hybrid of the variational autoencoder and the generative adversarial network, IDVAE, which does not need an explicit discriminator network. IDVAE shares a common decoder and generator network, and partially shares the encoder and the discriminator network. The qualitative and quantitative experiments on real-world benchmark datasets demonstrates that IDVAE (and its variant IDVAE(R)) performs on par and sometimes better than the state of the art hybrid approaches. We also show that IDVAE can be easily extended to work in a conditional setting, and experimentally demonstrate its performance on complex datasets.
Further, our results present inadequacies of the Fr\'echet Inception Distance and suggests an ensemble of experts for evaluating the quality of the generations. This can be further explored to derive a measure that does not require a model that is pre-trained on data from a different domain as that of the training samples.
{\small
\bibliographystyle{ieee}
\bibliography{ms}
}
\end{document}